\documentclass[letterpaper,10pt,conference]{ieeeconf}
\IEEEoverridecommandlockouts
\makeatletter

\let\AND\@undefined
\makeatother

\makeatletter
\let\NAT@parse\undefined
\makeatother
\usepackage{graphics} %
\usepackage{epsfig} %
\usepackage{mathptmx} %
\usepackage{times} %
\usepackage{amsmath} %
\usepackage{amssymb}%
\usepackage{graphicx}%
\usepackage{amssymb, psfrag, booktabs}	%
\usepackage{pdfpages}%
\usepackage{float}	%
\usepackage{subcaption} %
\usepackage{algpseudocode}
\usepackage{algorithm}

\usepackage{amsmath}
\usepackage{mathtools}
\usepackage{stfloats} %
\usepackage{pgfplots}
\usepackage{eurosym, color, xfrac}
\usepackage{placeins}
\usepackage{tikz}
\usetikzlibrary{shapes,arrows,arrows.meta,positioning,fit,backgrounds}
\usepackage{circuitikz}
\usepackage{bytefield}
\usetikzlibrary{decorations.pathreplacing}
\usepackage{adjustbox}
\usepackage{siunitx}
\usepackage{multirow}
\usepackage{fancyvrb}
\usepackage{xspace}
\usepackage{url}
\usepackage{pgfplots}
\usepackage{eurosym, color, xfrac}
\usepackage{placeins}
\usepackage{tikz}
\usetikzlibrary{shapes,arrows}
\usetikzlibrary{decorations.pathreplacing}
\usepackage{float}
\usepackage{relsize}
\usepackage{listings}
\usepackage{soul}
\usepackage{hyperref}
\hypersetup{
colorlinks=true,
linkcolor=blue,
citecolor = blue,
filecolor=black,
urlcolor=black,
}
\pgfplotsset{compat=1.18}

\title{\LARGE \bf
Optimization-Based Path Planning for Industrial Manipulators: Experimental Validation in Additive Manufacturing
}
\author{Shiva Ganesh Murali, Valentin Bhend, Deepak Ingole, Gabriele Fadini, Alisa Rupenyan
\thanks{ZHAW Centre for Artificial Intelligence, Zürich University of Applied Sciences,
Winterthur, 8400, Switzerland
{\tt\small \{\href{mailto:murv@zhaw.ch}{murv}, \href{mailto:bhed@zhaw.ch}{bhed}, \href{mailto:inge@zhaw.ch}{inge}, \href{mailto:fadi@zhaw.ch}{fadi},\href{rupn@zhaw.ch}{rupn\}@zhaw.ch}}
}
}
\begin{document}
\maketitle
\thispagestyle{empty}
\pagestyle{empty}
\begin{abstract}
\label{abstract}
Robotic additive manufacturing requires joint paths that reconcile robot feasibility with process-specific geometry. We develop an offline optimization-based path planner that combines joint-path regularity and self-collision penalties with process costs and constraints, so that a new process enters through its own reference, residuals, and tolerances. Full-path and boundary-coupled chunked optimization handle references of different lengths. Iterative Weight Tuning (IWT) determines the cost weights from the gaps between Key Performance Indicators (KPIs) and their targets. In carbon-fiber winding and robotic 3D printing, the planned paths meet the geometric tolerances and all KPI targets, and both paths are executed on a UR10e, without extrusion for printing, with joint-level tracking recorded during execution.
\end{abstract}
\section{Introduction}
\label{sec:intro}
In robotic manufacturing, the process determines how the robot must move the tool or workpiece~\cite{rupenyan2024automation}. In additive manufacturing, robot configurations must accommodate the process geometry and tolerances along an ordered reference while respecting joint limits, self-collision clearance, and path continuity~\cite{weingartshofer2023optimization}. The planner must therefore translate the ordered process reference into a joint-space path that respects these requirements.

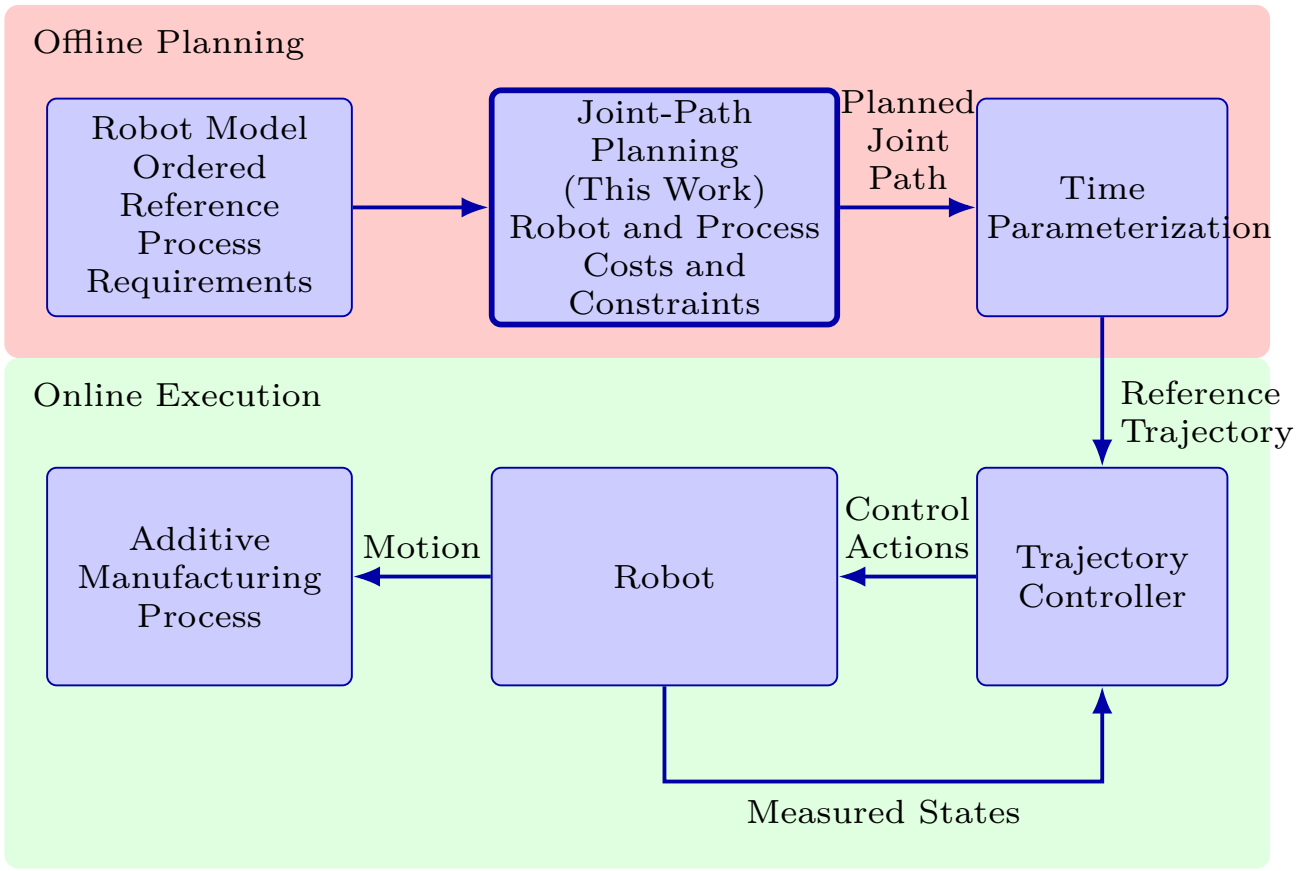
\begin{figure}[H]
\centering
\resizebox{\columnwidth}{!}{%
\begin{tikzpicture}[
font=\scriptsize,node distance=8mm,
block/.style={draw=blue!65!black,fill=blue!20,rounded corners=2pt,
align=center,text width=17mm,minimum height=16mm,inner sep=2pt},
leftblock/.style={block,text width=21mm},
middleblock/.style={block,text width=24mm},
flow/.style={-{Latex[length=2mm,width=1.5mm]},thick,draw=blue!65!black}]
\node[leftblock] (input) {Robot Model\\Ordered Reference\\Process Requirements};
\node[middleblock,right=10mm of input,very thick] (planner)
{\textbf{Joint-Path Planning}\\\textbf{(This Work)}\\Robot and Process\\Costs and Constraints};
\node[block,right=10mm of planner] (timing) {Time\\Parameterization};
\node[block,below=11mm of timing] (controller) {Trajectory\\Controller};
\node[middleblock] (robot) at (planner.center |- controller.center) {Robot};
\node[leftblock] (process) at (input.center |- controller.center)
{Additive\\Manufacturing\\Process};
\draw[flow] (input)--(planner);
\draw[flow] (planner)--node[above,align=center] {Planned\\Joint\\Path} (timing);
\draw[flow] (timing)--node[pos=0.65,right,align=left] {Reference\\Trajectory} (controller);
\draw[flow] (controller)--node[above,align=center] {Control \\Actions} (robot);
\draw[flow] (robot)--node[above,align=center] {Motion} (process);
\coordinate[below=7mm of robot] (feedback);
\draw[flow] (robot.south)--(feedback)-|
node[pos=0.25,below] (measurement) {Measured States} (controller.south);
\node[fit=(input)(planner)(timing),inner xsep=3mm,inner ysep=6mm]
(offline) {};
\coordinate (phaseboundary) at ($(timing.south)+(0,-3mm)$);
\coordinate (onlinebottom) at ($(measurement.south)+(0,-2mm)$);
\begin{pgfonlayer}{background}
\path[fill=red!20,draw=none,rounded corners=3pt]
(offline.west |- phaseboundary) rectangle (offline.north east);
\path[fill=green!12,draw=none,rounded corners=3pt]
(offline.west |- onlinebottom) rectangle (offline.east |- phaseboundary);
\end{pgfonlayer}
\node[anchor=north west,font=\scriptsize\bfseries,inner xsep=2mm,inner ysep=0.5mm,yshift=-1.2mm]
at (offline.north west) {Offline Planning};
\node[anchor=north west,font=\scriptsize\bfseries,inner xsep=2mm,inner ysep=0.5mm,yshift=-1.2mm]
at (offline.west |- phaseboundary) {Online Execution};
\end{tikzpicture}}
\caption{Offline planning and online execution pipeline, highlighting the joint-path optimization addressed in this work.}
\label{fig:manufacturing_control_architecture}
\end{figure}

We address the offline path-planning problem highlighted in Fig.~\ref{fig:manufacturing_control_architecture}. The planner selects joint configurations along the ordered process reference using robot and process costs and constraints. Time parameterization then provides a reference for online tracking with joint-state feedback.

Sampling-based methods search admissible configurations and connections on constraint manifolds~\cite{kingston2018constraints,orthey2024samplingreview}. IKLink links inverse-kinematic solutions along prescribed end-effector references~\cite{wang2024iklink}. Winding additionally requires core-relative fiber directions and fiber--robot clearance models. Printing requires nozzle-pose tolerances along the reference.

Optimization-based planners express costs and constraints. Covariant Hamiltonian Optimization for Motion Planning (CHOMP) uses covariant gradients, Stochastic Trajectory Optimization for Motion Planning (STOMP) uses stochastic perturbations, and TrajOpt uses sequential convex optimization~\cite{zucker2013chomp,kalakrishnan2011stomp,schulman2014trajopt}. cuRobo combines parallel inverse kinematics, geometric planning, and trajectory optimization~\cite{sundaralingam2023curobo}. Their kinematic and collision-handling tools require application-specific reference construction and process residuals for winding and printing. Local optimization remains sensitive to initialization.

Once a path is fixed, execution introduces timing and feedback requirements. Optimal control and Model Predictive Control (MPC) address these requirements with custom trajectory-optimization and MPC formulations \cite{mower2023optas} and with robust tube MPC which handles model uncertainty through  corridor-based planning \cite{wullt2026robust}. Receding-horizon optimization supports reactive motion within its horizon, while coordinating configurations over the complete manufacturing reference would require the same process geometry together with execution models and timing assumptions.

Learned models offer another way to generate candidate paths or initialize numerical optimization. Motion Policy Networks learn reactive motion from depth observations~\cite{fishman2023mpinets}, while DiffusionSeeder and Planning with Environment Representation, Sampling, and Trajectory Optimization (PRESTO) generate candidate paths for numerical refinement~\cite{huang2025diffusionseeder,seo2025presto}. Reinforcement Learning (RL) has also been combined with conventional planning and neural inverse kinematics for redundant manipulators~\cite{li2022drlframework}, and with inverse-Jacobian control for simulated deburring~\cite{rahul2024deburring}. At the execution stage, data-driven trajectory generation and compensation have targeted residual vibration and tracking error in pick-and-place tasks~\cite{wu2026autogeneration}. For winding and printing, a learned planner would need training tasks and inputs or rewards that represent the process geometry. Its paths would also need to be checked against the relevant tolerances.

Manufacturing planners incorporate process tolerances~\cite{weingartshofer2023optimization}, optimize orientation and timing~\cite{chen2025cooptimization}, and enforce coupled-system constraints~\cite{ye2026pathconstrained}. Graph-based methods address winding sequences~\cite{mersch2025automation}. The question here is how core-relative fiber references and prescribed nozzle poses can each be converted into joint paths while coupling process agreement, path regularity, and clearance over the ordered samples. 
Pointwise inverse kinematics does not couple these path-wide requirements, so we integrate their residuals and constraints in a common robot-stage optimization. This coupling introduces one weight per cost term. The weights trade process agreement against regularity and clearance, while acceptance is decided by process tolerances, so manual tuning must be repeated for each process. Constrained Bayesian optimization has tuned the parameters of a contouring planner jointly with low-level controller gains against motion performance metrics \cite{rupenyan2026data}, and NMPC weights have been tuned with Bayesian optimization \cite{fadini2026bayesian} and adapted across repeated winding tasks \cite{ingole2026iterativetuning}. We therefore pair the formulation with Iterative Weight Tuning (IWT), which determines the weights from the gaps between the evaluated metrics and their tolerances. The contributions of this work are as follows:
\begin{enumerate}
\item A process-aware joint-path formulation that couples winding fiber alignment or printing nozzle poses with joint regularity, self-clearance, and joint bounds. Both applications use forward kinematics to evaluate their process residuals within a common robot-stage optimization.
\item An adaptation of our Iterative Weight Tuning (IWT) method~\cite{ingole2026iterativetuning} to select robot and process cost weights for both winding and printing. Empirical sensitivities relate weight changes to robot and process Key Performance Indicator (KPI) errors, enabling regularized norm-optimal updates without differentiating through the path optimizer. The tuning uses clearance and fiber-alignment KPIs for winding and position- and orientation-tracking KPIs for printing.
\item A two-stage winding planner that converts groove transitions into a sequence of contact points and fiber directions in the core frame. The robot stage uses that fixed sequence to plan the motion of the core while accounting for fiber--robot clearance and enforcing an angular tolerance at groove endpoints.
\item A printing implementation that limits the number of active configurations per optimization. Fixed neighboring configurations and joint-difference penalties couple adjacent chunks, and alternating forward and reverse sweeps propagate the boundary updates.
\end{enumerate}

Section~\ref{sec:problem} presents the path-planning formulation and weight-tuning procedure.
\section{OPTIMIZATION-BASED PATH PLANNING}
\label{sec:problem}
\subsection{Problem Statement}
\label{sec:general_problem_statement}
Given a robot model, workpiece geometry $\mathcal W$, and a fixed, ordered process reference $\mathcal P=(r_1,\ldots,r_N)$, the offline planner must select one joint configuration per reference sample. Each $r_i$ specifies the process geometry to be realized at sample $i$. Configurations must satisfy joint bounds and any hard process constraints. The objective balances reference agreement, joint-path regularity, and self-clearance. Time parameterization and feedback tracking follow path planning.

Let $n_q$ denote the number of robot joints. At sample $i$ of the ordered path, the joint configuration vector is $\mathbf Q_i=[Q_{1,i},\ldots,Q_{n_q,i}]^\top\in\mathbb R^{n_q}$, where $Q_{j,i}$ is the position of joint $j$ at sample $i$ and $(\cdot)^\top$ denotes transpose. All vectors are columns, and $\|\cdot\|$ and $\|\cdot\|_2$ denote the Euclidean norm. For $N$ reference samples, the complete joint-space path is the decision variable
\begin{equation}
\bar{\mathbf Q}=[\mathbf Q_1,\ldots,\mathbf Q_N]\in\mathbb R^{n_q\times N}.
\label{eq:path_variables}
\end{equation}
The planner selects the complete joint path in~\eqref{eq:path_variables} so that the cost can account for differences between neighboring configurations. The superscript $(0)$ denotes initialization, and $\star$ denotes an optimized solution.

Separating robot-dependent terms from process-dependent terms gives the general optimization problem as follows:
\begin{subequations}
\label{eq:abstract_path_planning_problem}
\begin{align}
 \underset{\bar{\mathbf Q}}{\operatorname{minimize}}\quad&
 J_{\mathrm{robot}}(\bar{\mathbf Q})
 +J_{\mathrm{proc}}(\bar{\mathbf Q};\mathcal W,\mathcal P)
 \label{eq:abstract_planning_objective}\\
 \text{subject to}\quad&
 \mathbf h_{\mathrm{robot}}(\bar{\mathbf Q})=\mathbf0,
 \label{eq:abstract_robot_equalities}\\
 &\mathbf g_{\mathrm{robot}}(\bar{\mathbf Q})\leq\mathbf0,
 \label{eq:abstract_robot_inequalities}\\
 &\mathbf h_{\mathrm{proc}}(\bar{\mathbf Q};\mathcal W,\mathcal P)=\mathbf0,
 \label{eq:abstract_process_equalities}\\
 &\mathbf g_{\mathrm{proc}}(\bar{\mathbf Q};\mathcal W,\mathcal P)\leq\mathbf0.
 \label{eq:abstract_process_inequalities}
\end{align}
\end{subequations}
In~\eqref{eq:abstract_path_planning_problem}, the robot cost $J_{\mathrm{robot}}$ accounts for joint-path regularity and self-clearance. The process cost $J_{\mathrm{proc}}$ measures agreement with $\mathcal P$ and includes any process clearance penalties. For $\sigma\in\{\mathrm{robot},\mathrm{proc}\}$, the vectors $\mathbf h_\sigma\in\mathbb R^{m_{\mathrm{eq},\sigma}}$ and $\mathbf g_\sigma\in\mathbb R^{m_{\mathrm{ineq},\sigma}}$ collect equality and componentwise inequality constraints, respectively. The nonnegative integers $m_{\mathrm{eq},\sigma}$ and $m_{\mathrm{ineq},\sigma}$ count their scalar constraints. A zero count denotes an empty vector.

For each evaluated process metric $\mu$, the error $E_\mu(\bar{\mathbf Q};\mathcal P)$ must satisfy its tolerance $\varepsilon_\mu$ for the path to be accepted. Winding imposes its edge-end angular limit in~\eqref{eq:winding_edge_alignment}. The printing pose tolerances and penalized clearances are evaluated after optimization using the checks in~\eqref{eq:path_acceptance_criteria}.

\subsection{Path-Planning Formulation}
\label{sec:path_planning_formulation}
Forward kinematics maps $Q \in \mathbb{R}^{n_q}$ to the pose of the robot-attached tool or workpiece frame $E$ in the base frame $B$:
\begin{equation}
{}^{B}\mathbf T_E(\mathbf Q)=
\begin{bmatrix}
{}^{B}\mathbf R_E(\mathbf Q)&{}^{B}\mathbf p_E(\mathbf Q)\\
\mathbf 0^\top&1
\end{bmatrix}\in\mathrm{SE}(3),
\label{eq:planning_forward_kinematics}
\end{equation}
where ${}^{B}R_{E}(Q) \in SO(3)$ is the rotation and ${}^{B}p_{E}(Q) \in \mathbb{R}^{3}$ the translation. Here $\mathrm{SO}(3)$ denotes the special orthogonal group.

Equation~\eqref{eq:general_path_planning_problem} evaluates the robot and process costs on the same joint path. The vectors $\dot{\mathbf Q}_i,\ddot{\mathbf Q}_i,\dddot{\mathbf Q}_i\in\mathbb R^{n_q}$ are backward differences with respect to the unit-spaced path index, defined in~\eqref{eq:planning_path_velocity}--\eqref{eq:planning_path_jerk}. They are derived from the sole robot-stage decision variable $\bar{\mathbf Q}$. Execution timing is specified separately. In~\eqref{eq:planning_objective}, $w_v$, $w_a$, and $w_j$ weight the squared norms of the first, second, and third joint-path differences, respectively. The weight $w_{\mathrm{col}}$ scales the self-clearance penalty. All cost weights in this paper are nonnegative real scalars. The set $\mathcal C$ contains the checked robot-link pairs. For pair $c\in\mathcal C$, the penalty is active when its signed surface clearance $d_c(\mathbf Q_i)$ falls below the requested minimum $d_{\mathrm{safe},c}$. Signed distances are positive for separation, zero at contact, and negative for penetration.

The vectors $\mathbf Q_{\min},\mathbf Q_{\max}\in\mathbb R^{n_q}$ give the lower and upper joint-position bounds in~\eqref{eq:planning_joint_bounds}. For these robot constraints, $m_{\mathrm{eq},\mathrm{robot}}=0$ and $m_{\mathrm{ineq},\mathrm{robot}}=2n_qN$. The process model supplies the cost $J_{\mathrm{proc}}$ and any constraints in~\eqref{eq:planning_process_equalities}--\eqref{eq:planning_process_inequalities}.

\begin{figure*}[!b]
\centering
\begingroup
\vspace{-5mm}
\begin{subequations}
\label{eq:general_path_planning_problem}
\begin{align}
\underset{\bar{\mathbf Q}}{\operatorname{minimize}}\quad&
\underbrace{\sum_{i=1}^{N}\Biggl(
    \underbrace{w_v\|\dot{\mathbf Q}_i\|_2^2}_{\text{velocity cost}}
    +\underbrace{w_a\|\ddot{\mathbf Q}_i\|_2^2}_{\text{acceleration cost}}
    +\underbrace{w_j\|\dddot{\mathbf Q}_i\|_2^2}_{\text{jerk cost}}
    +\underbrace{w_{\mathrm{col}}\sum_{c\in\mathcal C}
        \bigl[\max\bigl(0,d_{\mathrm{safe},c}-d_c(\mathbf Q_i)\bigr)\bigr]^2}_{\text{self-collision penalty}}
\Biggr)}_{\text{robot costs}}
+\underbrace{J_{\mathrm{proc}}(\bar{\mathbf Q};\mathcal W,\mathcal P)}_{\text{process cost}}\label{eq:planning_objective}\\
\text{subject to}\quad&
\makebox[34mm][l]{$\displaystyle \mathbf Q_{\min}\leq\mathbf Q_i\leq\mathbf Q_{\max}$}\makebox[59mm][l]{$\displaystyle \forall i=1,\ldots,N$}\quad\mathrel{\tikz[baseline=-0.5ex]{\draw[line width=0.45pt,-{Latex[length=1.6mm,width=1.3mm]}] (0,0)--(-5mm,0);}}\quad\text{robot inequality constraints}\label{eq:planning_joint_bounds}\\
&\makebox[34mm][l]{$\displaystyle \dot{\mathbf Q}_i=\mathbf Q_i-\mathbf Q_{i-1}$}\makebox[26mm][l]{$\displaystyle \forall i=2,\ldots,N$}\makebox[33mm][l]{$\displaystyle \dot{\mathbf Q}_1=\mathbf 0$}\quad\mathrel{\tikz[baseline=-0.5ex]{\draw[line width=0.45pt,-{Latex[length=1.6mm,width=1.3mm]}] (0,0)--(-5mm,0);}}\quad\text{first joint-path difference}\label{eq:planning_path_velocity}\\
&\makebox[34mm][l]{$\displaystyle \ddot{\mathbf Q}_i=\dot{\mathbf Q}_i-\dot{\mathbf Q}_{i-1}$}\makebox[26mm][l]{$\displaystyle \forall i=3,\ldots,N$}\makebox[13mm][l]{$\displaystyle \ddot{\mathbf Q}_i=\mathbf 0$}\makebox[20mm][l]{$\displaystyle \forall i=1,2$}\quad\mathrel{\tikz[baseline=-0.5ex]{\draw[line width=0.45pt,-{Latex[length=1.6mm,width=1.3mm]}] (0,0)--(-5mm,0);}}\quad\text{second joint-path difference}\label{eq:planning_path_acceleration}\\
&\makebox[34mm][l]{$\displaystyle \dddot{\mathbf Q}_i=\ddot{\mathbf Q}_i-\ddot{\mathbf Q}_{i-1}$}\makebox[26mm][l]{$\displaystyle \forall i=4,\ldots,N$}\makebox[13mm][l]{$\displaystyle \dddot{\mathbf Q}_i=\mathbf 0$}\makebox[20mm][l]{$\displaystyle \forall i=1,2,3$}\quad\mathrel{\tikz[baseline=-0.5ex]{\draw[line width=0.45pt,-{Latex[length=1.6mm,width=1.3mm]}] (0,0)--(-5mm,0);}}\quad\text{third joint-path difference}\label{eq:planning_path_jerk}\\
&\makebox[93mm][l]{$\displaystyle \mathbf h_{\mathrm{proc}}(\bar{\mathbf Q};\mathcal W,\mathcal P)=\mathbf0$}\quad\mathrel{\tikz[baseline=-0.5ex]{\draw[line width=0.45pt,-{Latex[length=1.6mm,width=1.3mm]}] (0,0)--(-5mm,0);}}\quad\text{process equality constraints}\label{eq:planning_process_equalities}\\
&\makebox[93mm][l]{$\displaystyle \mathbf g_{\mathrm{proc}}(\bar{\mathbf Q};\mathcal W,\mathcal P)\leq\mathbf0$}\quad\mathrel{\tikz[baseline=-0.5ex]{\draw[line width=0.45pt,-{Latex[length=1.6mm,width=1.3mm]}] (0,0)--(-5mm,0);}}\quad\text{process inequality constraints}.\label{eq:planning_process_inequalities}
\end{align}
\end{subequations}
\endgroup
\end{figure*}

\subsection{Planning Procedure}
\label{sec:method}
Algorithm~\ref{alg:offline_planning} starts from the workpiece geometry $\mathcal W$, robot model $\mathcal R$, and an optional supplied geometric path $\mathcal P_{\mathrm{in}}$. It prepares the process reference $\mathcal P$ and an initial joint path $\bar{\mathbf Q}^{(0)}$ with matching sample indices. Optimization returns a candidate $\bar{\mathbf Q}^{\star}$. The validation report records whether the candidate meets the joint bounds, process tolerances, and clearance requirements.
\begin{algorithm}
\caption{Offline Joint-Path Optimization}
\label{alg:offline_planning}
\begin{algorithmic}[1]
\Require Workpiece $\mathcal{W}$, robot model $\mathcal{R}$, optional path $\mathcal P_{\mathrm{in}}$
\Ensure Optimized joint path $\bar{\mathbf Q}^{\star}$, validation report
\State Prepare the process reference $\mathcal P$ using~\eqref{eq:process_reference_preparation}
\State Initialize the joint path $\bar{\mathbf Q}^{(0)}$ using~\eqref{eq:initial_path}
\State Assemble and solve~\eqref{eq:general_path_planning_problem} from $\bar{\mathbf Q}^{(0)}$
\State Check feasibility and validate the path using~\eqref{eq:path_acceptance_criteria}
\State \Return $\bar{\mathbf Q}^{\star}$ and validation report
\end{algorithmic}
\end{algorithm}

\subsubsection{Process Reference Preparation}
\label{sec:method_reference}
Process preparation expresses the task as an ordered reference. It constructs the reference from the workpiece geometry or samples a supplied geometric path:
\begin{equation}
\mathcal P=\mathcal A(\mathcal W,\mathcal P_{\mathrm{in}})
=(r_1,\ldots,r_N).
\label{eq:process_reference_preparation}
\end{equation}
The map $\mathcal A$ collects the application-specific preparation steps, including sequencing and any separate process optimization. Winding uses $\mathcal P_{\mathrm{in}}=\varnothing$ and constructs the reference from $\mathcal W$. Printing samples the supplied pose path $\mathcal P_{\mathrm{in}}$. Once prepared, $\mathcal P$ stays fixed during the joint-path solve.

To evaluate a reference point in the robot base frame $B$, we transform its workpiece-frame coordinates ${}^{W}\mathbf p_i^{\mathrm{ref}}$ as follows:
\begin{equation}
{}^{B}\mathbf p_i^{\mathrm{ref}}
={}^{B}\mathbf R_W\,{}^{W}\mathbf p_i^{\mathrm{ref}}+{}^{B}\mathbf p_W.
\label{eq:reference_frame_transform}
\end{equation}
In~\eqref{eq:reference_frame_transform}, the rotation ${}^{B}\mathbf R_W$ maps vectors from frame $W$ to frame $B$. The vector ${}^{B}\mathbf p_W$ locates the origin of $W$ in $B$. The rigid transform is constant for a stationary workpiece and depends on the joint configuration for a robot-carried workpiece. The superscript $\mathrm{ref}$ identifies a reference point. Fiber directions use the rotation alone.
\subsubsection{Joint-Path Initialization}
\label{sec:method_initialization}
The numerical solve starts from
\begin{equation}
\bar{\mathbf Q}^{(0)}=[\mathbf Q_1^{(0)},\ldots,\mathbf Q_N^{(0)}],
\quad \mathbf Q_{\min}\leq\mathbf Q_i^{(0)}\leq\mathbf Q_{\max},\;\forall i.
\label{eq:initial_path}
\end{equation}
We obtain the initial joint path by applying inverse kinematics along the reference or repeating the first configuration. This path supplies a starting point for optimization. In winding, it also determines the initial contact point used by the soft anchor in~\eqref{eq:winding_robot_objective}.
\subsubsection{Cost and Constraint Construction}
\label{sec:method_terms}
At each candidate configuration, forward kinematics in~\eqref{eq:planning_forward_kinematics} gives the pose of the attached tool or workpiece. The process model uses that pose to evaluate fiber alignment for winding or nozzle pose error for printing. The robot terms in~\eqref{eq:planning_objective} evaluate joint-path differences and link clearance.

Joint bounds and specified alignment inequalities are posed as hard constraints, whereas self-clearance and fiber--robot clearance are soft penalties. Acceptance uses~\eqref{eq:path_acceptance_criteria}, with IWT also checking clearance against the fixed floor and feeder column.
\subsubsection{Joint-Path Problem Assembly}
\label{sec:method_problem}
The zero entries in~\eqref{eq:planning_path_velocity}--\eqref{eq:planning_path_jerk} omit terms with insufficient preceding samples from the common sum in~\eqref{eq:planning_objective}.

Interpolated checks, when enabled, use the configuration $\mathbf Q_{i,\xi}=(1-\xi)\mathbf Q_i+\xi\mathbf Q_{i+1}$, where $\xi\in(0,1)$ is the interpolation fraction and $i=1,\ldots,N-1$. These configurations are derived from adjacent path columns. Winding IWT validation used two interior configurations per interval.

Both applications use the joint bounds in~\eqref{eq:planning_joint_bounds}. Winding additionally imposes the edge-end alignment bound in~\eqref{eq:winding_edge_alignment}.
\subsubsection{Numerical Solution}
\label{sec:method_solution}
Full-path mode updates all $Nn_q$ joint coordinates together from $\bar{\mathbf Q}^{(0)}$. The solver uses automatic differentiation to compute objective gradients. The returned candidate $\bar{\mathbf Q}^\star$ can depend on the initial path.

Chunk mode updates a contiguous subset of configurations at a time. Neighboring configurations outside that subset are held fixed during the update. Boundary terms couple the active samples to those neighbors. Section~\ref{sec:printing} specifies how the printing implementation passes boundary information between chunks.
\subsubsection{Path Validation}
\label{sec:method_validation}
Validation compares process-error measures $E_\mu$ with tolerances $\varepsilon_\mu$ and checks clearances:
\begin{equation}
\begin{aligned}
E_\mu(\bar{\mathbf Q}^\star;\mathcal P)&\leq\varepsilon_\mu,
&&\mu\in\mathcal M_{\mathrm{val}},\\
d_c^{\mathrm{val}}(\mathbf Q)&\geq d_{\mathrm{acc},c},
&&(\mathbf Q,c)\in\mathcal V_{\mathrm{col}}.
\end{aligned}
\label{eq:path_acceptance_criteria}
\end{equation}
The set $\mathcal M_{\mathrm{val}}$ indexes the process metrics and their evaluation domains. The set $\mathcal V_{\mathrm{col}}$ contains the configuration--pair combinations checked for clearance. For each pair $c$, $d_c^{\mathrm{val}}$ denotes its signed surface clearance and $d_{\mathrm{acc},c}$ its acceptance threshold. For self-collision pairs, $d_c^{\mathrm{val}}=d_c$. For fiber--robot pairs, $d_c^{\mathrm{val}}=d_c^f$. Environment pairs use their corresponding signed surface gaps. The IWT acceptance thresholds are the category targets in Table~\ref{tab:iwt_progression}(a). Validation also checks the joint bounds in~\eqref{eq:planning_joint_bounds} and any active process constraints in~\eqref{eq:planning_process_equalities}--\eqref{eq:planning_process_inequalities}. These tests verify tolerances and clearances at the path samples and at the interpolated configurations.

\subsection{Iterative Weight Tuning}
\label{sec:iwt_method}
\begingroup\color{black}
Selecting the weights in~\eqref{eq:planning_objective} requires balancing competing costs to satisfy process-level KPIs and robot performance requirements. These tuning criteria differ from the planning objective and include nondifferentiable minimum-clearance and maximum-error metrics. A derivative-based approach would require differentiating the planning solution map through its optimality conditions, generally involving second-order derivatives of the objective and constraints and any regularization terms. This can be expensive and numerically sensitive when active constraints or local solutions change. We therefore adapt IWT~\cite{ingole2026iterativetuning}, which uses empirical sensitivities and requires no differentiation of the optimizer.

Let $\mathbf w\in\mathbb R_{>0}^{n_w}$ contain $n_w$ tunable weights. For each of $n_K$ KPIs, $K_a$ and $K_a^{\mathrm{tar}}>0$ denote its value and target. Define $e_a=(K_a-K_a^{\mathrm{tar}})/K_a^{\mathrm{tar}}$ for upper bounds and its negative for lower bounds; positive gaps indicate violations. The transformed gaps $\widetilde e_a=\operatorname{sgn}(e_a)\ln(1+|e_a|)$ and logarithmic weights $\theta_\ell=\log_{10}w_\ell$ form the vectors $\widetilde{\mathbf e}$ and $\boldsymbol\theta$, respectively, where $\operatorname{sgn}$ denotes the sign function.

Each offline repetition solves the planning problem and evaluates its KPIs. Changes between repetitions, denoted by $\Delta$, update a prior-initialized sensitivity estimate $\widehat{\mathbf S}\in\mathbb R^{n_K\times n_w}$ through the empirical model $\Delta\widetilde{\mathbf e}\approx\widehat{\mathbf S}\Delta\boldsymbol\theta$. The regularized norm-optimal update prioritizes violated targets within weight and step bounds. The resulting path is checked against~\eqref{eq:path_acceptance_criteria}.
\endgroup

\section{Application-Specific Formulations}
\label{sec:usecases}
Winding and printing share the robot costs and joint bounds in~\eqref{eq:general_path_planning_problem}. The applications differ in how the process reference is obtained and which geometric quantities the planned path must realize. During execution, the controller tracks the joint reference derived from that path.

\subsection{Robotic Carbon-Fiber Winding}
\label{sec:winding}
\begin{figure}
	\centering
	\begin{subfigure}[b]{0.48\columnwidth}
		\centering
		\includegraphics[trim=0 0 0 40bp,clip,height=1.10\linewidth]{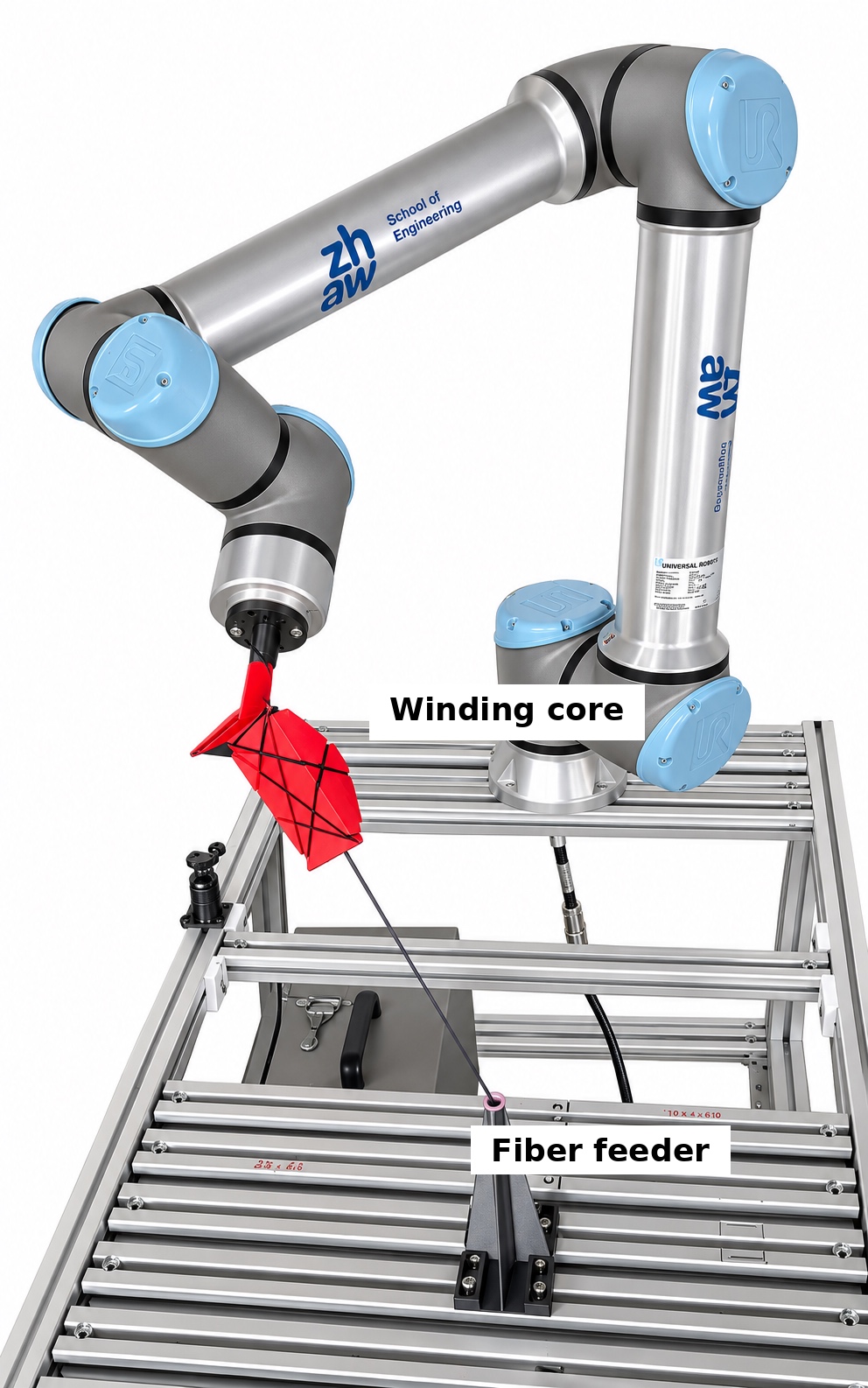}
		\caption{UR10e winding setup.}
		\label{fig:winding_setup_robot}
	\end{subfigure}
	\hfill
	\begin{subfigure}[b]{0.48\columnwidth}
		\centering
		\includegraphics[trim=0 8bp 0 24bp,clip,height=1.10\linewidth]{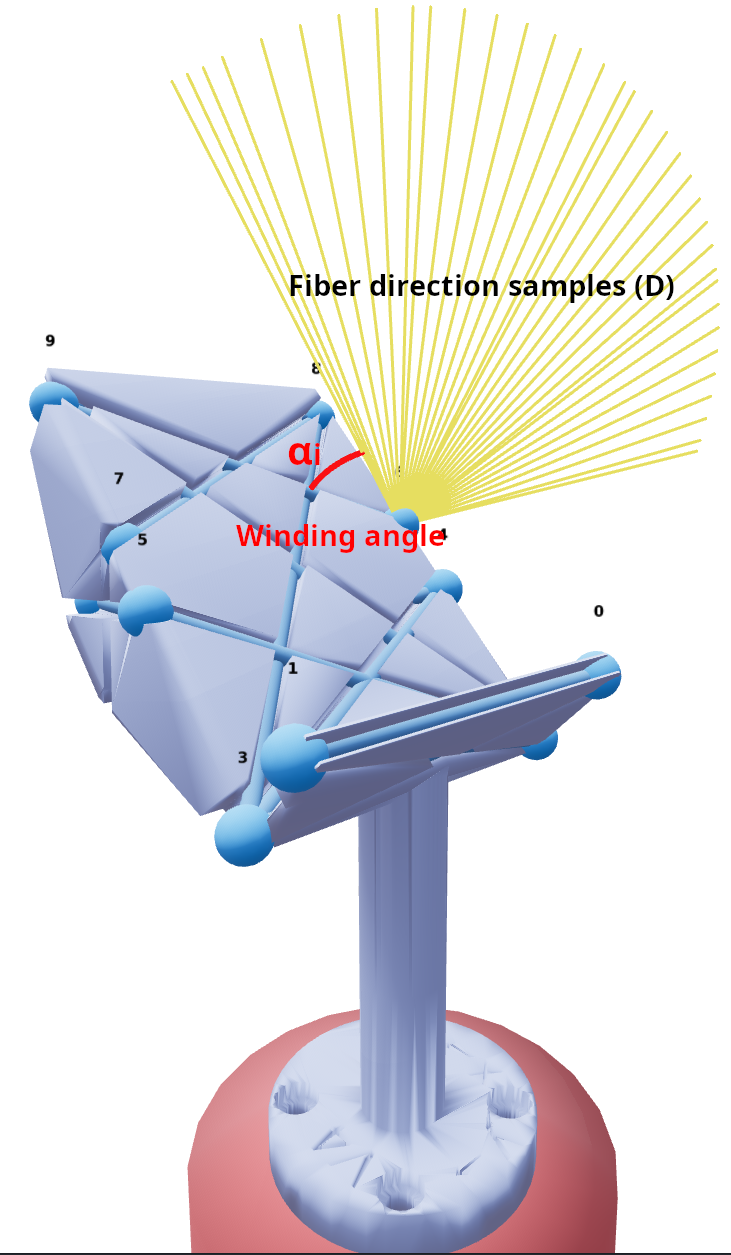}
		\caption{Gas pedal winding core.}
		\label{fig:winding_core}
	\end{subfigure}
	\caption{Experimental winding setup: (a) UR10e with the robot-mounted winding core and (b) winding core graph representation.}
	\label{fig:winding_setup}
\end{figure}

In winding, the robot carries a grooved core past a fixed fiber feeder (Fig.~\ref{fig:winding_setup}). The core pose determines how the free fiber approaches each groove. As the robot moves between grooves, it must align the fiber, keep it clear of the links, and respect joint limits and self-clearance.

The core vertices, groove edges, and edge normals define the winding geometry. Hierholzer\textquotesingle s algorithm first orders the vertices. Problem~1 in~\eqref{eq:winding_fiber_reference_problem} then computes the fiber directions for each groove transition. The resulting contact points and directions form the reference for Problem~2 in~\eqref{eq:general_path_planning_problem}, with the winding terms in~\eqref{eq:winding_process_cost}, to obtain a joint path that realizes it.
\subsubsection{Fiber-Reference Planning}
For one groove transition, let $M$ denote the number of fiber-direction samples and $\mathbf v\in\mathbb R^3$ the pivot vertex. The decision matrix $\mathbf D=[\mathbf d_1,\ldots,\mathbf d_M]\in\mathbb R^{3\times M}$ collects the corresponding unit fiber-direction vectors $\mathbf d_k\in\mathbb R^3$, expressed in the core frame. Problem~1 selects these directions as follows.

\begin{subequations}
\label{eq:winding_fiber_reference_problem}
\begin{align}
&\min_{\mathbf D}\;
\underbrace{w_{\mathrm s}\sum_{k=2}^{M}\|\mathbf d_k-\mathbf d_{k-1}\|_2^2}_{\text{fiber-direction regularity}}
+\underbrace{w_{\mathrm{plane}}\sum_{k=2}^{M-1}
[\max(0,\mathbf d_k^\top\mathbf n)]^2}_{\text{plane-side penalty}}
\label{eq:winding_fiber_objective}\\
&\text{subject to}\notag\\
&\makebox[35mm][l]{$\displaystyle \mathbf d_1=\mathbf d_{\mathrm{in}}$}\makebox[20mm][l]{$\displaystyle $}\;\mathrel{\tikz[baseline=-0.5ex]{\draw[line width=0.45pt,-{Latex[length=1.3mm,width=1mm]}] (0,0)--(-3mm,0);}}\;\text{incoming}\label{eq:winding_incoming_direction}\\
&\makebox[35mm][l]{$\displaystyle \mathbf d_M=\mathbf d_{\mathrm{out}}$}\makebox[20mm][l]{$\displaystyle $}\;\mathrel{\tikz[baseline=-0.5ex]{\draw[line width=0.45pt,-{Latex[length=1.3mm,width=1mm]}] (0,0)--(-3mm,0);}}\;\text{outgoing}\label{eq:winding_outgoing_direction}\\
&\makebox[35mm][l]{$\displaystyle \|\mathbf d_k\|_2=1$}\makebox[20mm][l]{$\displaystyle \forall k=1,\ldots,M$}\;\mathrel{\tikz[baseline=-0.5ex]{\draw[line width=0.45pt,-{Latex[length=1.3mm,width=1mm]}] (0,0)--(-3mm,0);}}\;\text{unit norm}\label{eq:winding_unit_directions}\\
&\makebox[35mm][l]{$\displaystyle \phi_b(\mathbf v+\tau_jL_f\mathbf d_k)\geq\rho_j^{\mathrm{fiber}}$}\makebox[20mm][l]{$\displaystyle \forall(k,j,b)\in\mathcal I_f$}\;\mathrel{\tikz[baseline=-0.5ex]{\draw[line width=0.45pt,-{Latex[length=1.3mm,width=1mm]}] (0,0)--(-3mm,0);}}\;\text{clearance}.\label{eq:winding_reference_clearance}
\end{align}
\end{subequations}

The boundary conditions in~\eqref{eq:winding_incoming_direction} and~\eqref{eq:winding_outgoing_direction} prescribe the incoming and outgoing directions $\mathbf d_{\mathrm{in}}$ and $\mathbf d_{\mathrm{out}}$. Equation~\eqref{eq:winding_unit_directions} requires each direction to have unit norm. In~\eqref{eq:winding_fiber_objective}, the nonnegative weights $w_{\mathrm s}$ and $w_{\mathrm{plane}}$ penalize changes in direction and deviations to the undesired side of the winding plane, respectively. The vector $\mathbf n\in\mathbb R^3$ is the unit normal to the incoming winding plane. Setting $w_{\mathrm{plane}}=0$ disables the plane-side penalty.

The clearance constraint in~\eqref{eq:winding_reference_clearance} uses the signed-distance functions $\phi_b:\mathbb R^3\rightarrow\mathbb R$ for the core and wrist, expressed in the core frame. The index $b$ identifies the checked body. The modeled fiber length is $L_f$. Sample $j$ lies at fraction $\tau_j=j/N_f$ along that length and requires clearance $\rho_j^{\mathrm{fiber}}$. Here $N_f$ is the number of fiber samples. The superscript $\mathrm{fiber}$ distinguishes this reference-stage surface-clearance bound from the centerline-separation threshold used in the robot stage. The constraint applies over $\mathcal I_f=\{1,\ldots,M\}\times\{1,\ldots,N_f\}\times\{\mathrm{core},\mathrm{wrist}\}$.

The optimized directions form $\mathbf D^\star=[\mathbf d_1^\star,\ldots,\mathbf d_M^\star]$. Each retained sample $k$ maps to a global index $i$ with $r_i=(\mathbf v_i,\mathbf d_i)=(\mathbf v,\mathbf d_k^\star)$. Concatenating these samples in winding order gives the fixed reference $\mathcal P$ of $N$ samples for the robot stage.

\subsubsection{Robot Joint-Path Optimization}
Problem~2 holds $\mathcal P$ fixed and optimizes all joint configurations in $\bar{\mathbf Q}$ together using~\eqref{eq:general_path_planning_problem}. The winding cost in~\eqref{eq:winding_robot_objective} uses the fiber geometry relative to the robot. Equation~\eqref{eq:winding_edge_alignment} adds an alignment bound at groove endpoints.
\begin{subequations}
\label{eq:winding_process_cost}
\begin{align}
&\begin{aligned}[t]
&J_{\mathrm{proc}}^{\mathrm{wind}}(\bar{\mathbf Q})=\underbrace{w_{\mathrm{align}}\sum_{i=1}^{N}
 [1-(\mathbf R_i\mathbf d_i)^\top\hat{\mathbf f}_i]^2}_{\text{fiber-direction alignment}}\\
 &+\underbrace{w_f\sum_{s\in\mathcal S}\sum_{c\in\mathcal F}
 \bigl[\max(0,d_{\mathrm{safe},c}^{f}-d_c^f(\mathbf Q_s))\bigr]^2}_{\text{fiber--robot collision penalty}}\\
 &+\underbrace{w_{\mathrm{side}}\sum_{(s,c)\in\mathcal D_f}
 \left[\max\left(0,\frac{\rho_c^{\mathrm{side}}-(\mathbf n^{(0)}_{sc})^\top
 \boldsymbol\ell_{sc}(\bar{\mathbf Q})}{L_c}\right)\right]^2}_{\text{initialized fiber-side preservation}}\\
 &+\underbrace{w_0\|\mathbf a_1-\mathbf a_1^{(0)}\|^2}_{\text{initial-contact anchoring}}
\end{aligned}\label{eq:winding_robot_objective}\\
&\text{subject to}\notag\\
&\makebox[34mm][l]{$\displaystyle (\mathbf R_i\mathbf d_i)^\top\hat{\mathbf f}_i\geq\cos(\alpha_{\max})$}\makebox[19mm][l]{$\displaystyle \forall i\in\mathcal I_{\mathrm{end}}$}\;\mathrel{\tikz[baseline=-0.5ex]{\draw[line width=0.45pt,-{Latex[length=1.3mm,width=1mm]}] (0,0)--(-3mm,0);}}\;\text{alignment}.\label{eq:winding_edge_alignment}
\end{align}
\end{subequations}
The attached frame $E$ is the core frame. Forward kinematics in~\eqref{eq:planning_forward_kinematics} gives the core rotation $\mathbf R_i={}^{B}\mathbf R_E(\mathbf Q_i)$ and translation $\mathbf t_i={}^{B}\mathbf p_E(\mathbf Q_i)\in\mathbb R^3$. The contact point in the base frame is $\mathbf a_i=\mathbf R_i\mathbf v_i+\mathbf t_i\in\mathbb R^3$, with $\mathbf v_i\in\mathbb R^3$ expressed in the core frame. The free fiber points from that contact to the fixed fiber feeder at $\mathbf p_{\mathrm{feed}}\in\mathbb R^3$, giving $\hat{\mathbf f}_i=(\mathbf p_{\mathrm{feed}}-\mathbf a_i)/\|\mathbf p_{\mathrm{feed}}-\mathbf a_i\|\in\mathbb R^3$. This definition requires $\|\mathbf p_{\mathrm{feed}}-\mathbf a_i\|>0$.

The weight $w_{\mathrm{align}}$ penalizes misalignment between the reference and free-fiber directions in the base frame. At the edge endpoints indexed by $\mathcal I_{\mathrm{end}}$, the direction error must stay within $\alpha_{\max}$. The weight $w_f$ penalizes fiber--robot clearance violations at knots and selected interpolated configurations $\mathbf Q_s\in\mathbb R^{n_q}$, indexed by $\mathcal S$. Each $\mathbf Q_s$ is either a path column or an interpolated configuration defined in Section~\ref{sec:method_problem}. For each checked fiber--robot pair $c\in\mathcal F$, $d_c^f$ is the signed surface clearance and $d_{\mathrm{safe},c}^f$ is the requested clearance.

The weight $w_{\mathrm{side}}$ penalizes departures from the initialized fiber side for sample--pair combinations $(s,c)\in\mathcal D_f$. The vector $\boldsymbol\ell_{sc}\in\mathbb R^3$ is the closest-point displacement between the fiber and robot centerlines. It points from the fiber centerline toward the robot centerline. The unit vector $\mathbf n^{(0)}_{sc}\in\mathbb R^3$ is its direction at initialization, held fixed during optimization. The required centerline separation $\rho_c^{\mathrm{side}}$ is the sum of the two primitive radii and the requested clearance for pair $c$. The length $L_c>0$ normalizes the signed-separation residual. Both $\rho_c^{\mathrm{side}}$ and $L_c$ have units of length. Finally, $w_0$ penalizes displacement of the initial contact from $\mathbf a_1^{(0)}$, its position at initialization.

\subsection{Robotic 3D Printing}
\label{sec:printing}
The printing robot carries an extrusion head above a stationary build surface. The joint path must reproduce the prescribed nozzle poses while limiting abrupt joint changes and maintaining self-clearance. Related work combines curved-layer path smoothing with feed-rate scheduling under joint kinematic constraints~\cite{xie2020pathsmoothing}. Extrusion-force feedback can regulate material flow during deposition~\cite{guidetti2024force}. Here, we address the geometric joint path; execution timing, extrusion control, and material dynamics remain outside the formulation.

The Tool Center Point (TCP) pose reference is supplied directly, so printing starts at the robot optimization stage (Problem~2). Reference preparation in~\eqref{eq:process_reference_preparation} samples the supplied poses. For printing, the superscript $\mathrm{des}$ identifies prescribed pose quantities. Each sample $r_i=(\mathbf p_i^{\mathrm{des}},\boldsymbol\eta_i^{\mathrm{des}})$ gives a desired TCP position $\mathbf p_i^{\mathrm{des}}\in\mathbb R^3$ and a unit orientation quaternion $\boldsymbol\eta_i^{\mathrm{des}}\in\mathbb R^4$ in the robot base frame. The robot costs and bounds remain as in~\eqref{eq:planning_objective}--\eqref{eq:planning_joint_bounds}. The printing cost is as follows:
\begin{equation}
J_{\mathrm{proc}}^{\mathrm{print}}(\bar{\mathbf Q})=
\sum_{i=1}^{N}\Bigl(
\underbrace{w_p\|\mathbf p_i-\mathbf p_i^{\mathrm{des}}\|^2}_{\text{position error penalty}}
+\underbrace{w_R[1-(\boldsymbol\eta_i^\top\boldsymbol\eta_i^{\mathrm{des}})^2]}_{\text{orientation error penalty}}
\Bigr).
\label{eq:printing_process_cost}
\end{equation}
The attached frame $E$ is the nozzle TCP frame. At sample $i$, $\mathbf p_i={}^{B}\mathbf p_E(\mathbf Q_i)\in\mathbb R^3$ is the TCP position. The unit quaternion $\boldsymbol\eta_i\in\mathbb R^4$ represents the rotation from~\eqref{eq:planning_forward_kinematics}. The nonnegative weights $w_p$ and $w_R$ penalize position and orientation error, respectively. The orientation term is invariant to quaternion sign. Setting $w_R=0$ gives position-only tracking. The process-constraint vectors in~\eqref{eq:planning_process_equalities}--\eqref{eq:planning_process_inequalities} are empty for printing. The pose tolerances are checked after optimization.

For long pose references, the method in Section~\ref{sec:method_solution} updates a contiguous subset of $\bar{\mathbf Q}$ at a time. Each update uses up to four fixed neighbors on either side. For difference order $h=1,\ldots,4$, a nonnegative boundary weight $\beta_h$ multiplies the squared Euclidean norm of each $h$th forward joint difference whose stencil crosses a chunk boundary. A stencil is the sequence of $h+1$ consecutive configurations used to compute that difference by repeated subtraction. The four boundary weights form $\boldsymbol\beta=[\beta_1,\ldots,\beta_4]^\top\in\mathbb R_{\geq0}^4$. These terms supplement the shared objective during chunk updates. Alternating forward and reverse sweeps pass boundary information between chunks. The position and orientation checks use the assembled path.

\section{Experimental Evaluation}
\label{sec:results}
We evaluated nominal path agreement for both applications and executed the planned paths on the physical winding and printing setups. Robot-reported joint targets and measured joint positions were recorded to assess execution tracking.
\subsection{Execution Setup}
\label{sec:execution_setup}
 Both the winding and printing paths were executed on a UR10e robot arm; the printing execution did not include extrusion. Printing used the UR10e kinematic model. Fig.~\ref{fig:r3pt_setup} illustrates the separate printing laboratory setup with an extrusion head. For winding execution, the planned joint path was converted into a time-indexed reference and streamed through the \texttt{servoJ} command in \texttt{ur\_rtde}, an interface based on Real-Time Data Exchange (RTDE)~\cite{lindvig2025urrtde}. The robot servo frequency was $500$\unit{\hertz}, corresponding to a control cycle of $2$\unit{\milli\second}. The robot's internal servo tracked the reference with a proportional gain of $300$ and a lookahead time of $0.2$\unit{\second}. Each \texttt{servoJ} call supplied speed and acceleration arguments of $0.5$\unit{\radian\per\second} and $0.5$\unit{\radian\per\second\squared}, respectively. Measured joint positions were logged to evaluate tracking.

Winding used $N=2401$ jointly optimized samples over $20$ edges. Hierholzer\textquotesingle s algorithm gave the one-based winding vertex sequence $[4,\allowbreak 1,\allowbreak 2,\allowbreak 3,\allowbreak 6,\allowbreak 16,\allowbreak 7,\allowbreak 10,\allowbreak 9,\allowbreak 8,\allowbreak 13,\allowbreak 5,\allowbreak 4,\allowbreak 11,\allowbreak 14,\allowbreak 10,\allowbreak 3,\allowbreak 12,\allowbreak 15,\allowbreak 9,\allowbreak 4]$. Printing used $N=479\,523$ poses in chunks of at most $90\,000$, with two alternating sweeps.

Offline path optimization ran on an NVIDIA H100 Graphics Processing Unit (GPU), with 64\unit{\giga\byte} memory. Both applications used JAXIPM~\cite{viljoen2026scaling} with a limit of $1000$ iterations per solve and tolerances of $10^{-6}$ for the Nonlinear Programming (NLP) solve and constraint residual. Angular residuals were evaluated in radians and distances in metres; reporting units are indicated in the figures and tables. Joint positions were bounded by $\pm360$\unit{\degree}. The joint order is shoulder pan, shoulder lift, elbow, and wrists~1--3. Table~\ref{tab:optimization_parameters} lists the geometric settings and acceptance tolerances.
\begin{table}
\centering
\begingroup\color{black}
\caption{Robot and process geometry settings, acceptance tolerances, and printing chunk-boundary weights.}
\label{tab:optimization_parameters}
\setlength{\tabcolsep}{1pt}
\begin{tabular}{@{}lc@{}}
\toprule
Parameter group & Value\\
\midrule
\multicolumn{2}{@{}l}{\textbf{Common robot settings}}\\
Self-clearance $d_{\mathrm{safe},c}$ (mm) & $30$\\
Numerical clearance tolerance (mm) & $0.030$\\
\midrule
\multicolumn{2}{@{}l}{\textbf{Winding parameters}}\\
$M,N_f$ & $(120,24)$\\
$\rho_1^{\mathrm{fiber}}$; $\rho_j^{\mathrm{fiber}}$, $j=2,\ldots,N_f$ (mm) & $(1,2)$\\
Fiber length $L_f$ (m) & $0.600$\\
Fiber radii: reference / robot stage (mm) & $(1,3)$\\
Fiber feeder position $\mathbf p_{\mathrm{feed}}$ (m) & $(-0.747,-0.014,0.223)$\\
Fiber--robot clearance $d_{\mathrm{safe},c}^{f}$ (mm) & $5$\\
Nominal edge-end limit $\alpha_{\max}$ ($^\circ$) & $2$\\
IWT all-knot / edge-end limits ($^\circ$) & $(2,0.5)$\\
\midrule
\multicolumn{2}{@{}l}{\textbf{Printing parameters}}\\
Boundary weights $\boldsymbol\beta$ & $(1,0.5,0.1,0.01)$\\
Position (mm), orientation ($^\circ$) tolerances & $(1,1)$\\
\bottomrule
\end{tabular}
\endgroup
\end{table}

\subsection{Robotic Carbon-Fiber Winding}
\label{sec:results_winding}
\begin{figure}
	\centering
    \includegraphics[width=\columnwidth]{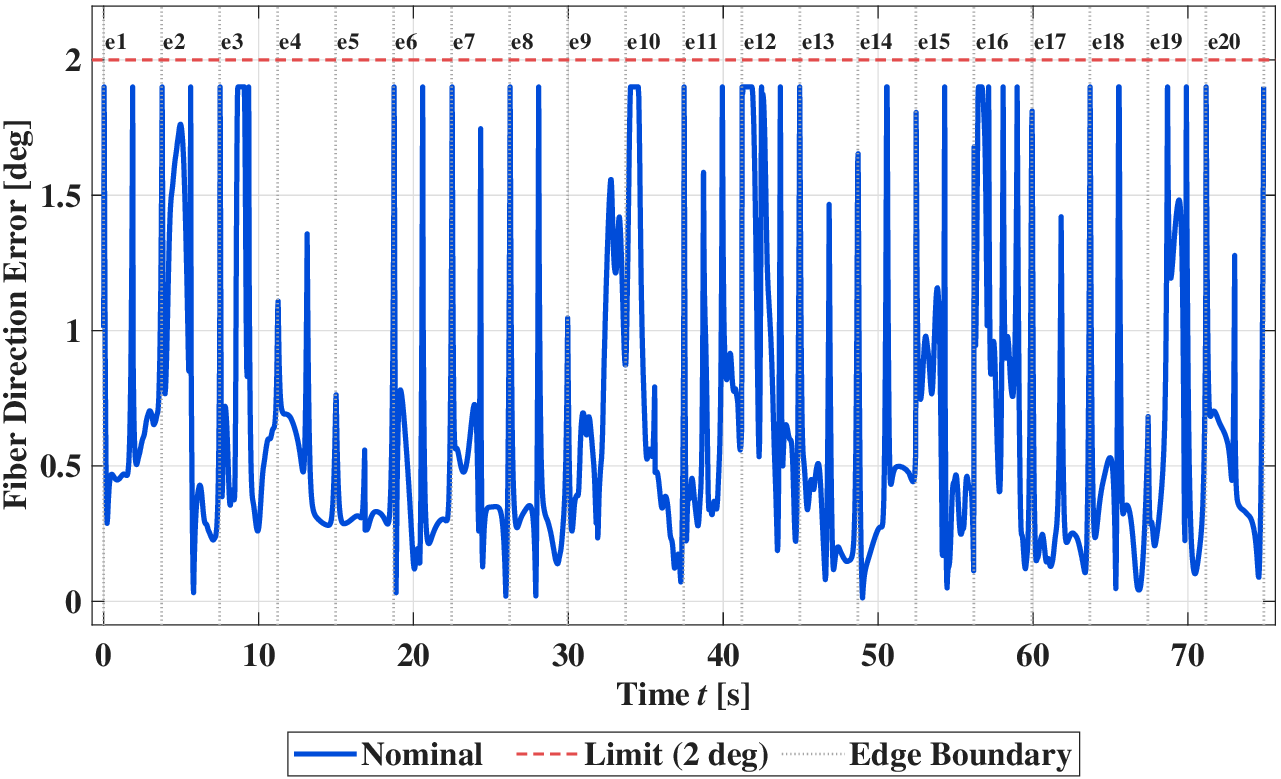}
	\caption{Nominal fiber-direction error during winding. Labels e1--e20 identify the edges, and vertical lines mark their boundaries.}
	\label{fig:winding_error}
\end{figure}

The first-stage winding weights were $(w_{\mathrm s},w_{\mathrm{plane}})=(1,10000)$. The second-stage winding weights were selected from repetition~5 in Table~\ref{tab:iwt_progression}(b). The side-preservation scale was $L_c=5$\unit{\milli\metre}, with a clearance allowance of $0.5$\unit{\milli\metre} in $\rho_c^{\mathrm{side}}$. For winding execution, JAXIPM computed an offline full-path solution using these selected weights. The prepared fiber reference remained fixed during this solve. Fig.~\ref{fig:winding_error} evaluates the final planned path through $\alpha_i=\arccos((\mathbf R_i\mathbf d_i)^\top\hat{\mathbf f}_i)$, the angle between the reference and free-fiber directions.

For the path in Fig.~\ref{fig:winding_error}, the nominal endpoint errors had a Root Mean Square (RMS) value of $1.236$\unit{\degree} and a maximum of $1.900$\unit{\degree}. The maximum was below the $2$\unit{\degree} groove-alignment tolerance, so every evaluated endpoint met the criterion. For the same planned path, the largest change of any joint between successive samples was $2.28$\unit{\degree}. Section~\ref{sec:iwt_results} reports separate winding IWT runs.

\begin{figure}
	\centering
	\includegraphics[width=\columnwidth]{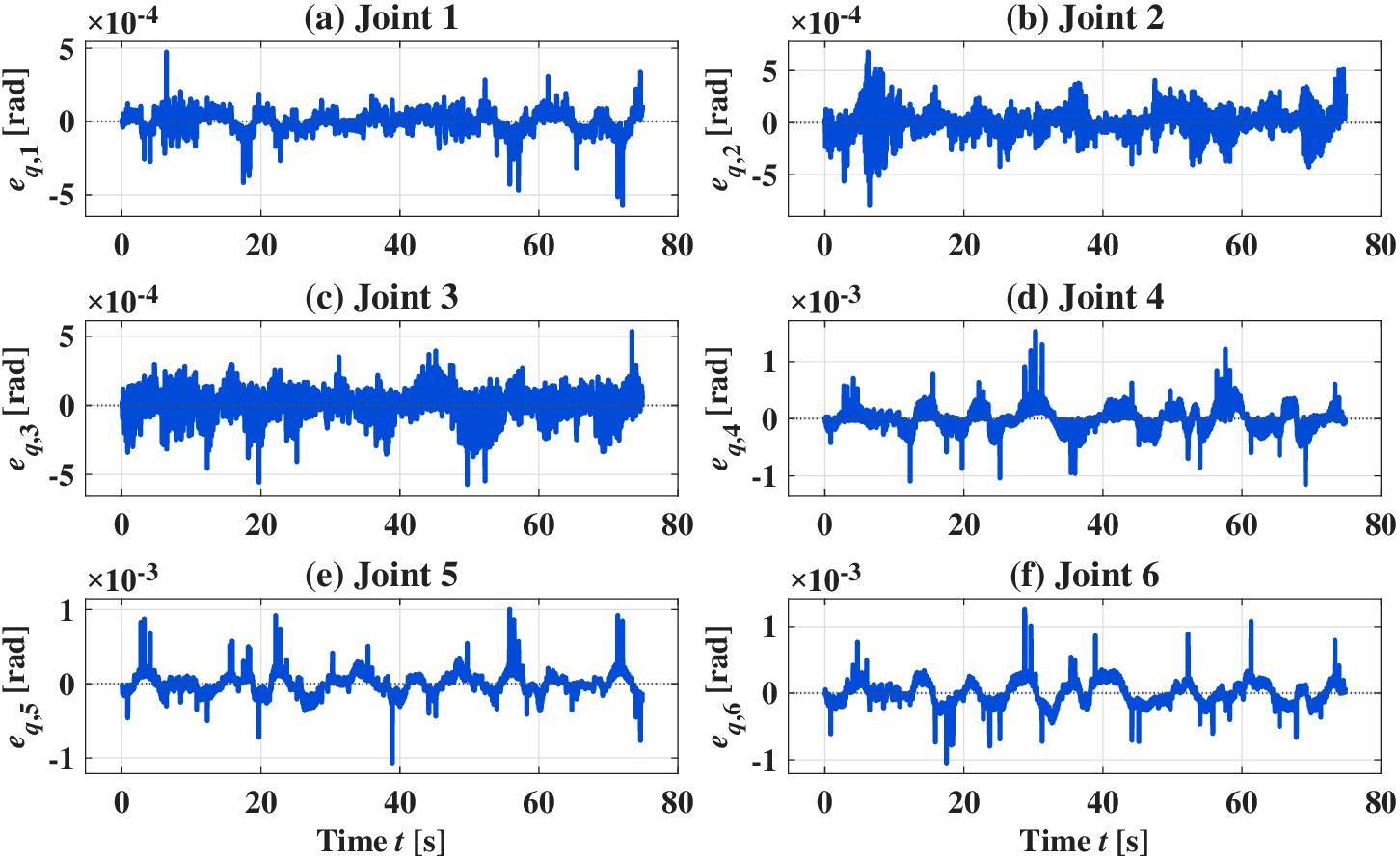}
	\caption{UR10e joint-position errors during winding, computed as measured position minus the robot-reported target for each of the six joints.}
	\label{fig:winding_measured_tracking}
\end{figure}

\paragraph{Measured execution}
Fig.~\ref{fig:winding_measured_tracking} presents experimental joint tracking recorded during physical winding on the UR10e. At execution time $t$, the error $e_{q,j}(t)=Q_j^{\mathrm{actual}}(t)-Q_j^{\mathrm{target}}(t)$ compares measured joint positions with the robot-controller targets returned by RTDE as \texttt{actual\_q} and \texttt{target\_q}. The planner path supplies the setpoints sent to \texttt{servoJ}. Fig.~\ref{fig:winding_error} separately evaluates nominal fiber alignment. The experiment validates joint-servo tracking during execution of the planned winding path.

\subsection{Path Planning for Robotic 3D Printing}
\label{sec:results_printing}
For the printing setup, IWT meets all three KPI targets after $3$ repetitions: maximum TCP position error below $1$\unit{\milli\metre}, maximum orientation error below $1$\unit{\degree}, and minimum robot self-clearance above $30$\unit{\milli\metre}. The repetition-$3$ weights are $(w_v,w_a,w_j,w_{\mathrm{col}})=(0.2358,0.2358,0.2358,1)$ and $(w_p,w_R)=(47.34,1.576)$. We optimize the chunks with alternating sweeps using~\eqref{eq:general_path_planning_problem} and the printing cost in~\eqref{eq:printing_process_cost}. Section~\ref{sec:printing} describes the boundary treatment. We evaluate the complete path after assembling the chunks, so the reported errors cover the entire sampled reference. The printing path was also executed on the physical setup, with joint targets and measured positions logged for tracking assessment (without extrusion).

\begin{figure}
    \centering
    \vspace{-2mm}
    \begin{subfigure}[b]{0.48\columnwidth}
        \centering
        \includegraphics[trim=0 0 0 8bp,clip,height=1.14\linewidth]{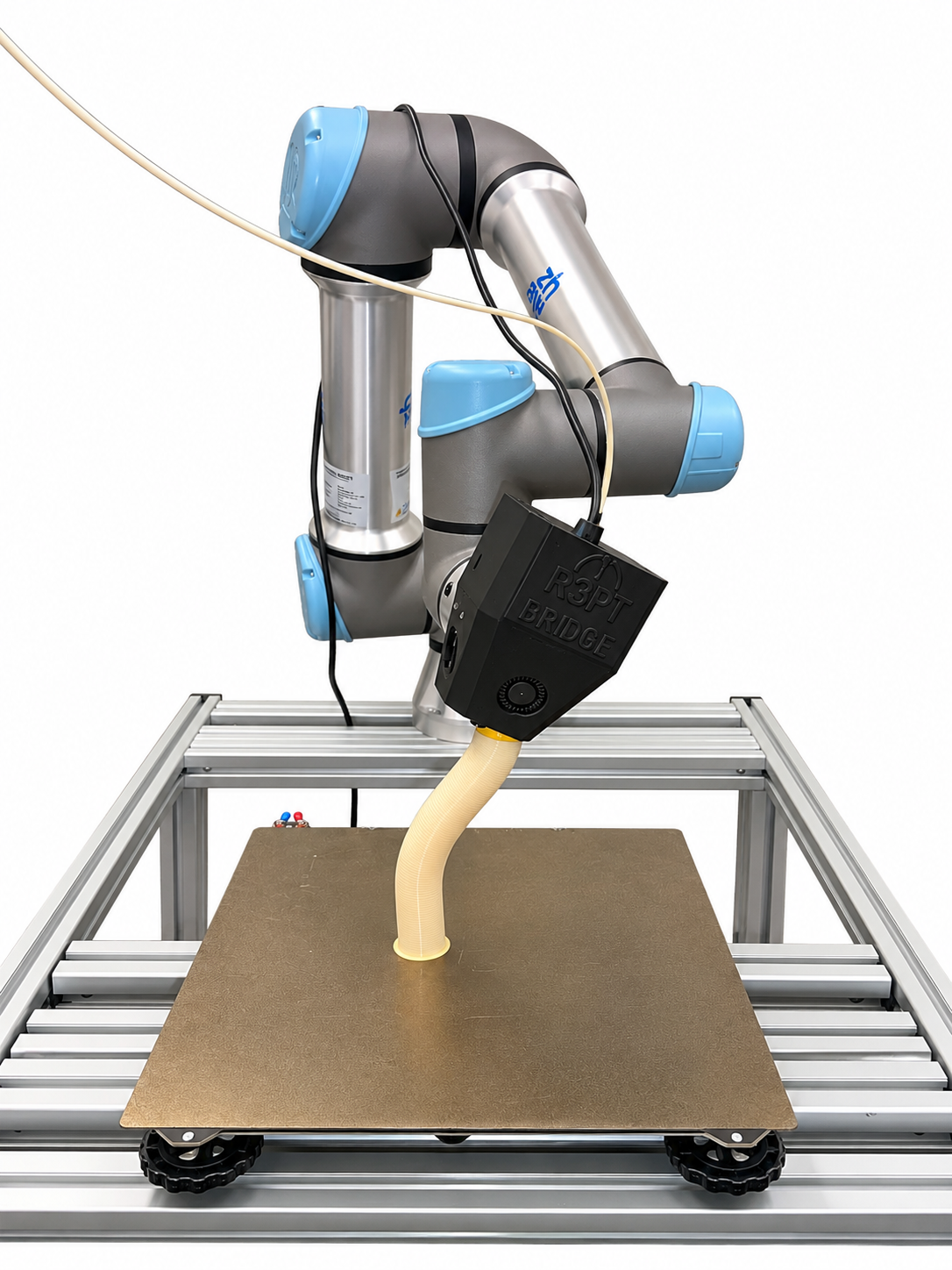}
        \caption{Printing lab setup.}
        \label{fig:r3pt_setup}
    \end{subfigure}\hfill
    \begin{subfigure}[b]{0.48\columnwidth}
        \centering
        \includegraphics[width=\linewidth,height=1.14\linewidth,keepaspectratio]{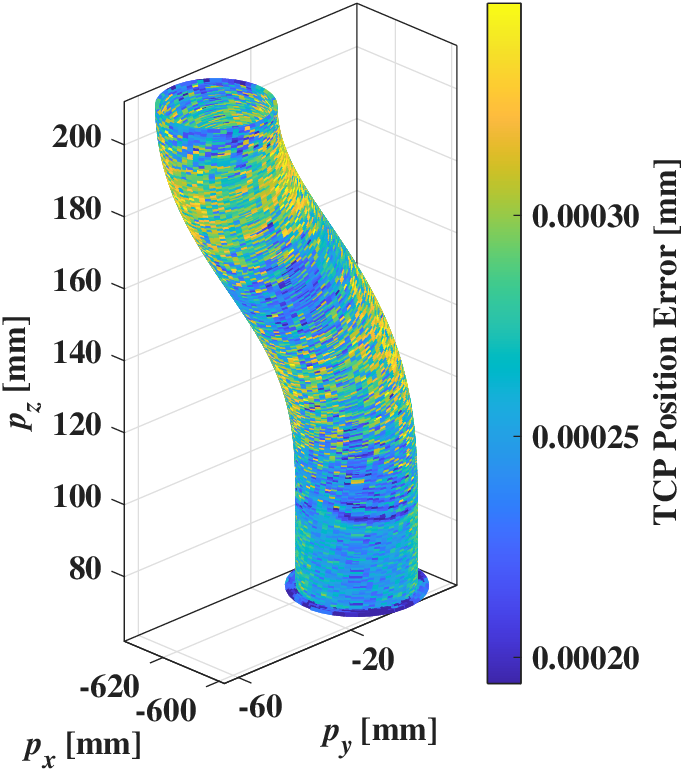}
        \caption{Path colored by TCP error.}
        \label{fig:printing_error}
    \end{subfigure}
    \caption{Robotic 3D printing: (a) UR10e setup with an extrusion head; (b) reference path colored by averaged nominal TCP position error. Reported maxima use all samples.}
    \label{fig:printing_setup_and_error}
\end{figure}

Fig.~\ref{fig:r3pt_setup} shows the printing setup with the robot-mounted extrusion head. Fig.~\ref{fig:printing_error} shows the reference path colored by averaged TCP position error between the reference and the forward kinematics of the planned joint path. Its axes $p_x,p_y,p_z$ denote Cartesian position components in the robot base frame; the statistics below use all path samples. The position error was $4.38\times10^{-4}$\unit{\milli\metre} RMS and $0.0660~$\unit{\milli\metre} at its maximum. The quaternion angle $2\arccos(|\boldsymbol\eta_i^\top\boldsymbol\eta_i^{\mathrm{des}}|)$ measures orientation error. It was $3.79\times10^{-5}$\unit{\degree} RMS and $1.32\times10^{-3}$\unit{\degree} at its maximum.

Both maxima are below the nominal acceptance tolerances of $1$\unit{\milli\metre} and $1$\unit{\degree}, respectively. The largest joint-position increment between successive samples is  $0.0187$\unit{\degree}.

\subsection{Iterative Weight Tuning}
\label{sec:iwt_results}
For winding IWT, the tuned vector is  $\mathbf w=(w_{\mathrm{col}},w_{\mathrm{align}},w_f,w_0,w_v,w_a,w_j,w_{\mathrm{side}})^\top$, with $n_w=8$ weights and $n_K=4$ KPIs. The corresponding bounds are  $w_{\mathrm{col}}\in[0.9,1100]$, $w_{\mathrm{align}}\in[0.9,11000]$, $w_f,w_0\in[0.09,11]$, and $w_v,w_a,w_j,w_{\mathrm{side}}\in[10^{-6},1]$. Each update changes each log-weight $\theta_\ell$ by at most one, so each weight changes by at most a factor of ten.

The winding study applies the IWT procedure in Section~\ref{sec:iwt_method} using the computational setup in Section~\ref{sec:execution_setup}. KPIs I, III, and IV report minimum signed clearances over the evaluated knots, interpolated configurations, and included pairs for robot--environment, robot--fiber, and self-clearance, respectively. KPI II reports the maximum fiber-direction error over all knots, with corresponding  targets listed in Table~\ref{tab:iwt_progression}(a).

Table~\ref{tab:iwt_progression} shows that at Repetition~1, the robot--fiber and self-clearance targets are met. By Repetition~3, environment clearance also meets its target, while the fiber-direction error remains above its limit. Repetition~5 meets all four KPI targets and passes the hard-constraint checks. Its maximum endpoint error is $0.498$\unit{\degree}, below the tightened endpoint limit, and the all-knot maximum in the table approaches the $2$\unit{\degree} bound. These results establish nominal sampled acceptance for the configured geometry. The final execution path is computed separately with the selected weights, as described in Section~\ref{sec:results_winding}.
\begin{table}
\centering
\begingroup\color{black}
\footnotesize
\caption{Winding KPI targets and results at selected IWT repetitions, with the corresponding weight progression.}
\label{tab:iwt_progression}
\setlength{\tabcolsep}{2pt}
\begin{tabular}{@{}lcrrr@{}}
\toprule
\multicolumn{5}{@{}l}{\textbf{(a) KPI targets and evaluated results}}\\
KPI & Target & Rep. 1 & Rep. 3 & Rep. 5\\
\midrule
KPI I: Environment clearance (mm) & $\geq5$ & $-4.677$ & $25.58$ & $23.82$\\
KPI II: Maximum fiber error ($^\circ$) & $\leq2$ & $6.296$ & $2.177$ & $1.998$\\
KPI III: Robot--fiber clearance (mm) & $\geq2$ & $7.585$ & $8.257$ & $7.525$\\
KPI IV: Self-clearance (mm) & $\geq25$ & $36.42$ & $36.51$ & $36.58$\\
KPIs met & 4 & 2 & 3 & 4\\
Hard checks & --- & No & No & Yes\\
\midrule
\multicolumn{5}{@{}l}{\textbf{(b) Weights obtained using IWT}}\\
$w_{\mathrm{col}}$ & --- & $1$ & $89.58$ & $58.35$\\
$w_{\mathrm{align}}$ & --- & $1$ & $1$ & $1.057$\\
$w_f,w_0$ (each) & --- & $1$ & $1$ & $1$\\
$w_v,w_a,w_j,w_{\mathrm{side}}$ (each) & --- & $1$ & $0.06052$ & $0.05883$\\
\bottomrule
\end{tabular}
\par\vspace{1mm}
\parbox{\columnwidth}{\scriptsize Rep. denotes repetition. Grouped weights have equal values. Hard checks cover joint bounds and the active alignment inequalities.}
\endgroup
\end{table}
The four regularity and side-preservation weights are tied and change together, while the fiber-clearance and anchor weights are held fixed, so IWT explores three independent directions with four update observations. The sensitivity model guides updates along these directions, and Table~\ref{tab:iwt_progression} shows selected KPI outcomes, including the initial and accepted paths. Nominal acceptance near an angular limit requires separate assessment of execution and calibration errors before physical process accuracy can be inferred.
\section{Conclusions}
\label{sec:concl}
The proposed offline planner uses a shared joint-path formulation for winding and printing, which accounts for robot self-clearance in both applications and for fiber--robot clearance in winding. The nominal winding path meets the groove-end alignment tolerance with a maximum error of $1.90$\unit{\degree}, and the assembled printing path meets its pose tolerances with maximum nominal errors of $0.066$\unit{\milli\metre} in position and $1.32\times10^{-3}$\unit{\degree} in orientation. IWT determines the weights of the path-planning problem from geometric KPIs for winding and from position and orientation KPIs for printing, and all targets are met in both applications. The physical execution on the UR10e validates joint-level tracking of the planned winding path, and measuring the placed fiber and the printed geometry is the next step. The offline formulation assumes a fixed feeder position, robot-base registration, and environment geometry, so changes to these quantities require updating the model and solving the optimization again. The formulation applies to manufacturing processes whose requirements can be expressed as costs and constraints on the joint path through forward kinematics. A new process of this class requires its own reference, cost terms, and tolerances, while the robot terms, the solver, and the tuning loop remain unchanged, and IWT then determines the weights that satisfy the process tolerances.

\section*{Acknowledgment}
This work was supported as a part of the National Centre of Competence in Research (NCCR) Automation, funded by the Swiss National Science Foundation (grant number 51NF40\textunderscore225155), and partially by Johann Jakob Rieter Stiftung and Rieter AG.
\bibliographystyle{IEEEtran}
\bibliography{bib/ieeeconf}
\end{document}